%% file: iclr2025_conference.tex
\documentclass{article} 
\usepackage{iclr2025_conference,times}

\input{math_commands.tex}

\usepackage{hyperref}
\usepackage{url}
\usepackage[utf8]{inputenc} 
\usepackage[T1]{fontenc}    
\usepackage{hyperref}       
\usepackage{url}            
\usepackage{booktabs}       
\usepackage{amsfonts}       
\usepackage{nicefrac}       
\usepackage{microtype}      
\usepackage{xcolor}         
\usepackage{natbib}
\usepackage{graphicx}
\usepackage{amsmath}
\usepackage{bbm}

\title{Investigating Memorization in Video Diffusion Models}

\author{
    Chen Chen\textsuperscript{1} \quad 
    Enhuai Liu\textsuperscript{1} \quad 
    Daochang Liu\textsuperscript{2} \quad 
    Mubarak Shah\textsuperscript{3} \quad 
    Chang Xu\textsuperscript{1} \\
    \textsuperscript{1}School of Computer Science, Faculty of Engineering, The University of Sydney, Australia \\
    \textsuperscript{2}School of Physics, Mathematics and Computing, The University of Western Australia, Australia \\
    \textsuperscript{3}Center for Research in Computer Vision, University of Central Florida, USA \\
    {\tt\small \{cche0711@uni., eliu0719@uni., c.xu@\}sydney.edu.au} \hspace{0.9mm} \\
    {\tt\small daochang.liu@uwa.edu.au} \hspace{0.9mm} 
    {\tt\small shah@crcv.ucf.edu}
}

\iclrfinalcopy 
\begin{document}

\maketitle

\input{sec/0_abstract}
\input{sec/1_intro}

\input{sec/3_metric}

\input{sec/4_case_study}
\input{sec/5_strategies}

\input{sec/6_conclusion}

\input{sec/acknowledgement}

\bibliography{iclr2025_conference}
\bibliographystyle{iclr2025_conference}

\appendix
\input{sec/X_suppl}

\end{document}

%% file: math_commands.tex
\usepackage{amsmath,amsfonts,bm}

\def\eqref#1{equation~\ref{#1}}

\def\1{\bm{1}}

\DeclareMathAlphabet{\mathsfit}{\encodingdefault}{\sfdefault}{m}{sl}
\SetMathAlphabet{\mathsfit}{bold}{\encodingdefault}{\sfdefault}{bx}{n}



%% file: sec/0_abstract.tex
\begin{abstract}
\vspace{-0.3cm}
Diffusion models, widely used for image and video generation, face a significant limitation: the risk of memorizing and reproducing training data during inference, potentially generating unauthorized copyrighted content. While prior research has focused on image diffusion models (IDMs), video diffusion models (VDMs) remain underexplored. To address this gap, we first formally define the two types of memorization in VDMs (content memorization and motion memorization) in a practical way that focuses on privacy preservation and applies to all generation types. We then introduce new metrics specifically designed to separately assess content and motion memorization in VDMs. Additionally, we curate a dataset of text prompts that are most prone to triggering memorization when used as conditioning in VDMs. By leveraging these prompts, we generate diverse videos from various open-source VDMs, successfully extracting numerous training videos from each tested model. Through the application of our proposed metrics, we systematically analyze memorization across various pretrained VDMs, including text-conditional and unconditional models, on a variety of datasets. Our comprehensive study reveals that memorization is widespread across all tested VDMs, indicating that VDMs can also memorize image training data in addition to video datasets. Finally, we propose efficient and effective detection strategies for both content and motion memorization, offering a foundational approach for improving privacy in VDMs. 
\end{abstract}

%% file: sec/1_intro.tex
\vspace{-0.6cm}
\section{Introduction}
\label{intro}
\vspace{-0.3cm}
Following the surge in popularity of image diffusion models (IDMs), which surpassed GANs~\citep{GAN} in image synthesis quality~\citep{iddpm_2021_icml}, video diffusion models (VDMs)~\citep{sora, gen3} have emerged as a significant advancement and attracted a lot of attention in both the research community and public due to their ability to generate novel and realistic videos.
However, a critical limitation has been identified in IDMs that raises questions about the novelty of their outputs, where training images can be extracted by a pretrained IDM and regurgitated during generation~\citep{carlini_2023_usenix}. Such memorized outputs have been criticized as ``digital forgery”\citep{somepalli_2023_cvpr}, and several lawsuits\citep{lawsuit} have been filed against companies for generating copyrighted artworks with these models.

Significant research has been conducted to understand and mitigate this memorization issue, identifying causes~\citep{somepalli_2023_neurips} and proposing remedies~\citep{ambient_diffusion, wen_2024_iclr, chen-amg, ren2024unveiling, chen_be, chen_prss}. These efforts have achieved notable success in IDMs. However, there is a lack of thorough investigation into memorization in VDMs, particularly in the influential text-conditional (T2V) models.
The investigation of memorization in VDMs is arguably even more critical than in IDMs, as videos consist of multiple temporally consistent frames. This adds an additional dimension to consider, and alongside content memorization, motion memorization also needs to be addressed. 
Motion patterns are highly distinct, much like biometrics, and their memorization raises significant privacy concerns. Previous work~\citep{motion_motivation_1, motion_motivation_2, motion_motivation_3} has shown that unique physical traits, such as a person’s gait, are sufficient for identification. Additionally, recent studies~\citep{motion_motivation_4, motion_motivation_5} demonstrate that motion data can be used to identify individuals in virtual reality (VR) environments. As a result, addressing motion memorization is just as crucial as tackling content memorization, especially given the growing use of VDMs and their broad range of downstream applications.
A recent study~\citep{frame_by_familiar_frame} attempted to explore memorization in VDMs. 
Despite their notable contribution, their definitions and evaluations of both content and motion memorization are narrow in scope and lack broad applicability, which has hindered their wider adoption in the research community.

\begin{figure}[tb]
  \centering
  \includegraphics[width=\linewidth]{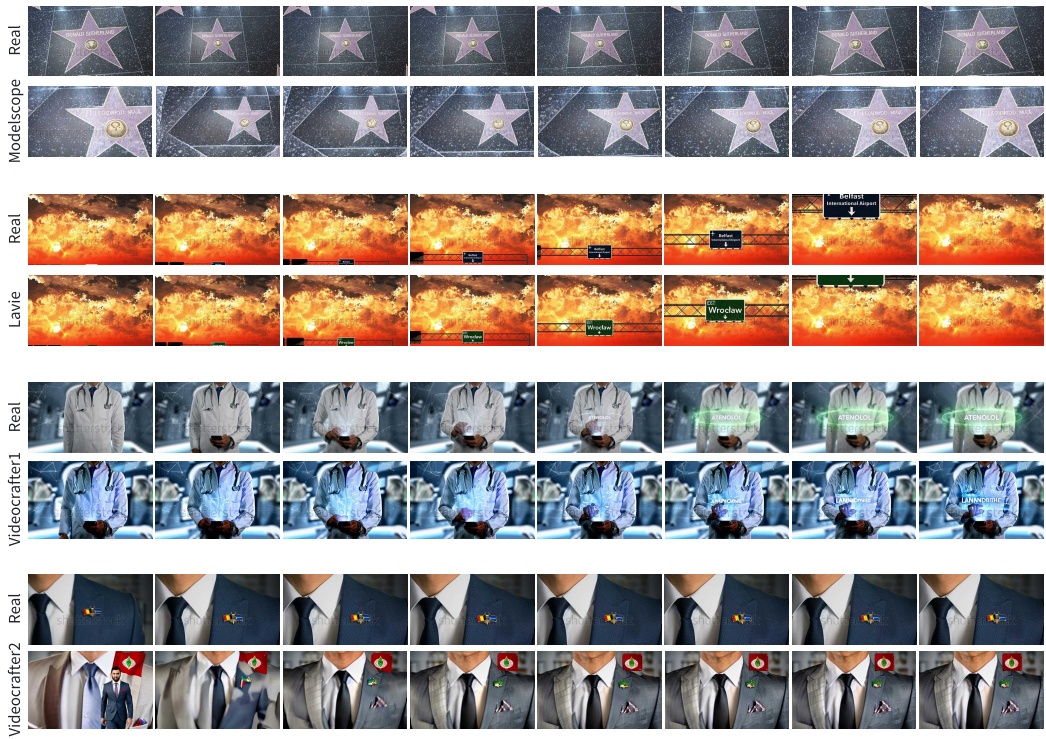}
  \vspace{-0.7cm}
  \caption{Video training dataset (WebVid-10M) being extracted by several open-source text-conditional VDMs. The generated videos can memorize both content and motion of the training videos. For each example, the top shows the training video and the bottom shows the generated video.}
  \label{fig:figure1}
\vspace{-0.6cm}
\end{figure}

\vspace{-0.2cm}
To address these limitations, we first redefine content memorization as a frame-level phenomenon independent of motion memorization in Sec. \ref{sec:metric}, addressing cases where the existing video-level definition fails, especially in content-only memorization scenarios. For motion memorization, we introduce a disentangled definition that applies to all generation types, overcoming the limitations of prior definitions that are entangled with content and only applicable to image-to-video (I2V) generation.
Then, we correspondingly develop new evaluation metrics tailored separately for content and motion memorization. Our content memorization metric shows improved accuracy over baseline, while our motion memorization metric is the first of its kind designed for VDMs. It emphasizes detecting motion patterns that pose privacy risks and has demonstrated reasonable performance in experiments.
Then, in Sec. \ref{sec:case_study}, using our proposed metrics and insights from IDM memorization research~\citep{somepalli_2023_neurips,wen_2024_iclr}, we are the first to systematically extract training data from various VDMs~\citep{modelscope,lavie,videocrafter1,videocrafter2,ramvid}, including text-conditional models trained on large datasets like WebVid-10M~\citep{webvid-10m} and obtain valuable findings. Unlike prior research that relied on individual instances of generated videos extracted from project websites, we provide a comprehensive analysis of memorization frequency by quantifying similarity scores and memorized generation percentages using our proposed metrics.
Our findings reveal that memorization is widespread, affecting not only video data but also image data used for training text-to-image (T2I) models, which serve as backbones for VDMs, introducing privacy risks. This aspect has been overlooked by previous research~\citep{frame_by_familiar_frame}, which even suggests T2I backbones as a potential remedy for the issue. Moreover, we found that VDMs pretrained on public datasets~\citep{webvid-10m,ucf-101} and fine-tuned on larger, higher-quality private datasets still memorize public data. This suggests that the overall extent of memorization may be underestimated, as these models may also memorize private datasets.
Finally, in Sec. \ref{sec:strategies}, we propose remedies by adapting image memorization detection strategies for VDMs. By incorporating the temporal dimension, we develop methods capable of detecting both content and motion memorization separately with high effectiveness and efficiency.

In summary, our contributions are five-fold: 
(1) We propose more practical, disentangled definitions for both content and motion memorization with a focus on privacy preservation.
(2) We develop new evaluation metrics tailored separately for content and motion memorization.
(3) By curating a dataset that contains 500 text prompts prone to memorization in VDMs, we are the first to systematically extract training data from various VDMs, including the under-explored text-conditional VDMs.
(4) We conduct a comprehensive analysis of memorization frequency by quantifying similarity scores and memorized generation percentages using our proposed metrics and reveal interesting findings.
(5) We propose the first strategies that enable both content motion memorization detection in VDMs, offering a strong foundation for future improvements.

%% file: sec/3_metric.tex
\section{Defining and measuring video memorization}
\label{sec:metric}
Previous work~\citep{frame_by_familiar_frame} highlights the need to analyze motion memorization in video diffusion models (VDMs) in addition to content memorization, which has been well-studied in image diffusion models (IDMs). They proposed two memorization definitions: \textit{`content and motion' memorization} and \textit{`motion' memorization}, along with evaluation methods. However, we identify several limitations in their definitions and metrics that may hinder broader adoption. To address this, we propose more practical definitions and more generalized, disentangled metrics for evaluating memorization in VDMs.

\subsection{Content memorization}
\label{content_mem}
\subsubsection{Research gap}
\label{content_mem_gap}
Content memorization has been extensively studied in IDMs, where it is defined as `reconstructive memory', which refers to the reproduction of an object that appears identically in a training image, neglecting minor variations in appearance that could result from data augmentation, with allowances for transformations like shifting, scaling, or cropping~\citep{somepalli_2023_cvpr, chen-amg}. Self-Supervised Copy Detection (SSCD)~\citep{sscd} is the standard pre-trained model used to extract embeddings from images, and cosine similarity between SSCD embeddings quantifies content memorization.

For VDMs, the memorization definitions and metrics need to be adapted due to the additional temporal dimension in videos.
The only work that has investigated content memorization in VDMs~\citep{frame_by_familiar_frame}, however, does not separately define content memorization as a disentangled definition. Instead, it combines content and motion memorization into a single definition, where a generated video is considered memorized if every frame perfectly replicates a training video in both content and motion. 
They correspondingly measure such \textit{content and motion memorization} by adapting SSCD, where they compute SSCD embeddings for each frame, concatenate them, and then calculate the cosine similarity of the resulting vectors, termed Video SSCD (VSSCD):
\begin{align}
\text{VSSCD} = & \, \text{cosine\_similarity} \Big(
\text{concat} \big( \Phi_{SSCD}(G_i) \big)_{i=1}^{N_1}, \text{concat} \big( \Phi_{SSCD}(T_j) \big)_{j=1}^{N_2}
\Big)
\end{align}
where \(G_i\) represents the \(i\)-th frame of the generated video, which has \(N_1\) frames, and \(T_j\) represents the \(j\)-th frame of the training video, which has \(N_2\) frames. The function \(\Phi_{SSCD}(\cdot)\) extracts the SSCD embedding for a frame.

However, this approach has several limitations:
\textit{Firstly}, it has no definition or metric for content-only memorization, where content is memorized, but the motion is not. This narrows the range of possible use cases and leads to incomplete evaluations.
\textit{Secondly}, it only captures video-level memorization and misses the more practical frame-level risks, however, privacy concerns can arise even if a single frame of a generated video memorizes a training frame.
\textit{Thirdly}, it does not assess memorization on image training sets. However, many VDMs, like those built on IDMs such as Stable Diffusion, are pretrained on large-scale image datasets. This creates a risk of generated videos memorizing these image datasets. We argue for a more generalized approach that unifies memorization assessments across both video and image domains.
\textit{Lastly}, \cite{frame_by_familiar_frame} only demonstrates the effectiveness of VSSCD by showing limited true positive examples with high VSSCD scores. However, it has not been quantitatively evaluated using standard classification measures such as F1-score \& AUC. Our analysis reveals many failure cases, motivating the need for an improved metric.

\subsubsection{Proposed definition and metric}
\label{content_mem_ours}
To address these limitations, we redefine content memorization as a frame-level phenomenon independent of motion memorization. A generated video is considered memorized if any single frame replicates a training video frame or a training image (in the case of single-frame data), following the rationale of `reconstructive memory' defined in image memorization studies.

We modify the SSCD similarity measure for video memorization by calculating the frame-wise similarity for all pairs of frames between generated and training videos rather than concatenating embeddings. The maximum similarity score across all pairs is then used as the video similarity score, formalized as the Generalized SSCD (GSSCD):
\begin{equation}
\text{GSSCD} = \max_{1 \leq i \leq N_1, \, 1 \leq j \leq N_2} \text{SSCD}(G_i, T_j)
\label{eq:gsscd}
\end{equation}
where \(G_i\) represents the \(i\)-th frame of the generated video with \(N_1\) frames, and \(T_j\) represents the \(j\)-th frame of the training video with \(N_2\) frames. The function \(SSCD(\cdot, \cdot)\) computes the cosine similarity between the SSCD embedding of two frames. This metric generalizes to compute similarities between videos and images (when the number of frames is one), making it applicable to both domains.

\vspace{-0.5cm}
\begin{table}[h]
\centering
\scriptsize 
\caption{Comparison of VSSCD and GSSCD. Bold values indicate the better-performing results.}
\label{table:vsscd_gsscd}
\begin{tabular}{l | c c}
\hline
\hline
& AUC & F1 \\
\hline
VSSCD & 0.974 & 0.814 \\
GSSCD & \textbf{0.995} & \textbf{0.919} \\
\hline
\end{tabular}
\end{table}
\vspace{-0.5cm}

\subsubsection{Evaluations}
\label{eval_gsscd}
We also quantitatively evaluate the alignment of the metrics with human annotations. Since WebVid-10M serves as the training dataset for most VDMs, including ModelScope, we leveraged the most duplicated captions from WebVid-10M, which have a higher likelihood of triggering memorization, as text-conditioning for ModelScope. Using these prompts, we generated 1,000 videos, ensuring a mix of memorized and non-memorized cases. Human annotators were then asked to label these generated videos as either memorized or non-memorized, with access to the corresponding training videos and captions. These human annotations served as ground truth for computing classification metrics (AUC and F1-score) for both VSSCD and GSSCD. As shown in Tab. \ref{table:vsscd_gsscd}, GSSCD aligns more closely with human annotations, significantly outperforming VSSCD in both AUC and F1-score. This demonstrates GSSCD’s effectiveness as a metric for evaluating content memorization.

\begin{figure}[tb]
  \centering
  \includegraphics[width=\linewidth]{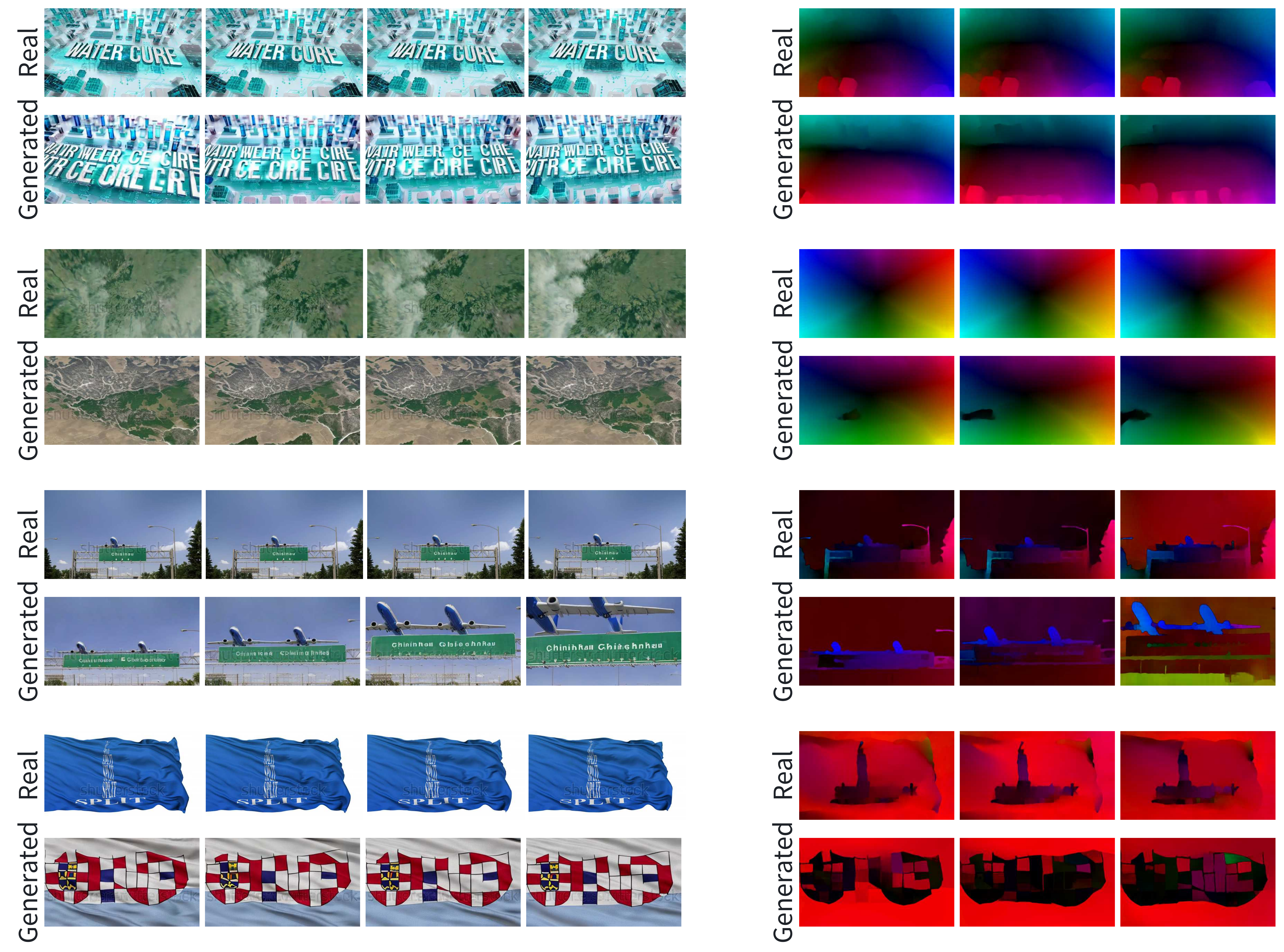}
  \caption{Motion memorization detected using OFS-3. The left side shows consecutive frame pairs from both the training video and the generated video using ModelScope, while the right side visualizes the optical flow between consecutive frames, computed using RAFT. OFS-3 calculates the average cosine similarity across three frame pairs, effectively capturing motion memorization. The OFS-3 scores for all four examples are above 0.5: 0.9012, 0.9393, 0.8523, and 0.8806, thus classified as motion-memorized cases.}
  \label{fig:figure5_positive}
\vspace{-0.5cm}
\end{figure}

\subsection{Motion memorization}
\label{motion_mem}
\subsubsection{Research gap}
\label{motion_mem_gap}
Motion memorization is not applicable for image generations and has not been extensively explored in video diffusion models (VDMs), and has only been indirectly studied by \cite{frame_by_familiar_frame}. Their approach focuses narrowly on image-conditional video generation (I2V), where they define motion memorization as the inability of the model to generate novel motion patterns when conditioned on augmented versions of an initial training video frame. The method's rationale is: a VDM capable of understanding motion should produce consistent high-fidelity videos when conditioned on both the original and augmented initial frames (e.g., flipping, cropping).

Instead of using a direct metric to evaluate motion memorization, they propose an indirect approach: comparing the fidelity of generated videos when conditioned on the original and varied versions of the initial frame. They found that conditioning on the original frame resulted in high-fidelity outputs due to perfect memorization of both content and motion, while conditioning on variants led to artifacts attributed to the model's memorization of motion patterns from the training videos.

However, this definition and evaluation have several limitations:
\textit{Firstly}, its applicability is limited in that it applies only to I2V scenarios and cannot be extended to other types of generations, such as unconditional or text-conditional.
\textit{Secondly}, the reliance on fidelity as a proxy for motion memorization introduces ambiguity, as high fidelity can also result from content memorization or other factors unrelated to motion. This approach fails to isolate the contribution of memorized motion patterns and to draw reliable conclusions about motion memorization.
\textit{Lastly}, it is resource-intensive, requiring numerous video generations to compute the FVD for fidelity assessment. The lack of a one-to-one similarity metric between individual generation-training pairs further reduces practicality and efficiency.

\vspace{-0.2cm}
\subsubsection{Proposed definition and metric}
\label{motion_mem_ours}
To address these shortcomings, we propose a more general and direct definition of motion memorization that applies to all types of video generations, not just I2V. Our approach evaluates motion alone, disentangling it from other elements like content or fidelity. We define \textbf{motion memorization} as the occurrence of high similarity in the optical flow of consecutive frames between a generated and a training video. Crucially, we disregard natural motions, such as camera panning or static frames, that pose no privacy risks.
To formalize this, we introduce a novel metric called \textbf{Optical Flow Similarity (OFS-k)}. 
Optical flow captures the motion trajectories within a video, abstracting away static content details and emphasizing dynamic movement patterns. This makes it ideal for comparing videos with different content, as it isolates motion signals from appearance.
Thus, OFS effectively quantifies the similarity between optical flows over consecutive frames, with \textbf{k} representing the number of consecutive frames considered in the computation. The key steps are elaborated as follows:

\textbf{Optical flow calculation}: For each generated-training video pair, we compute the optical flow for all consecutive frames using RAFT~\citep{raft}. For each flow pair \((\mathbf{f}^g, \mathbf{f}^t)\), we calculate the \textbf{cosine similarity} between the flow vectors at corresponding pixels:
\begin{equation}
S_{i, j} = \frac{\sum \mathbf{f}^g_i \cdot \mathbf{f}^t_j}{\|\mathbf{f}^g_i\| \|\mathbf{f}^t_i\|}
\label{eq:consine-ofs}
\end{equation}
    where \(\mathbf{f}^g_i\) and \(\mathbf{f}^t_j\) are the optical flow vectors for the \(i\)-th flow of generated video (from the \(i\)-th to \(i+1\)-th frame) and the \(j\)-th flow of training video (from the \(j\)-th to \(j+1\)-th frame). The summation and norm are computed over the x and y components of the optical flow vectors (the first dimension of size 2 in the flow tensor with shape \([2, H, W]\)), resulting in an output of shape \([H, W]\), where \(H\) and \(W\) represent the height and width of the frame. Finally, the cosine similarity \( S_{i, j} \) is averaged across all pixels to obtain the final similarity score \( \bar{S}_{i, j} \).

\textbf{Max consecutive flow similarity}: For a given \textbf{k}, we compute the cosine similarity for \textbf{k} consecutive frames and define \textbf{OFS-k} as the maximum average similarity score across all \textbf{k} consecutive frame windows. The rationale for considering consecutive frames is that memorized motion often spans continuous sequences of frames, and measuring the maximum over these sequences helps capture such behavior without being overly sensitive to isolated frame similarities. Formally, \textbf{OFS-k} is defined as:
\begin{equation}
\text{OFS-k} = \max_{0 \leq i \leq N_1 - k, \, 0 \leq j \leq N_2 - k} \left( \frac{1}{k} \sum_{n=0}^{k-1} \bar{S}_{i+n, j+n} \right)
\label{eq:ofs-k}
\end{equation}

\textbf{Hyperparameter \(k\)}: The hyperparameter \(k\) controls the length of consecutive frames considered. If \(k\) is too small, the metric may become too strict, potentially resulting in a high false positive (FP) rate by categorizing natural, non-memorized motions as memorized. Conversely, setting \(k\) too large could overlook cases where only a portion of the video is memorized, leading to false negatives (FN). Thus, \(k\) is a tunable parameter that can be adjusted to balance the trade-off between FP and FN.

\subsubsection{Natural Motion Filtering (NMF)}
\label{nmf}

To further refine our approach, we introduce \textbf{Natural Motion Filtering (NMF)} to filter out motions that should not be classified as memorized due to natural occurrence, targeting the two cases below.

\textbf{Camera panning}: This effect involves uniform motion across all pixels in the same direction. We compute the \textbf{entropy} of the flow directions to quantify the diversity of pixel movement. Low entropy indicates uniform motion, which we consider natural and not memorized. Specifically, we compute the angle \(\theta_i\) of each optical flow vector \(\mathbf{f}_i\) and create a histogram of these angles. The entropy \(H\) is then calculated as:
\begin{equation}
H = - \sum p(\theta_i) \log p(\theta_i)
\label{eq:entropy}
\end{equation}
where \(p(\theta_i)\) is the probability of a flow vector having direction \(\theta_i\). If \(H\) falls below a threshold, the motion is classified as camera panning and is ignored.
    
\textbf{Static frames}: Frames with minimal movement are considered static. We calculate the average magnitude of the optical flow vectors, and if the magnitude is below a specified threshold, the frame is categorized as static and excluded from further analysis.

By applying NMF, we ensure that only meaningful, privacy-sensitive motion patterns are evaluated as potential memorized motions, filtering out natural and non-threatening movements.

\begin{table}[t]
\caption{Comparison of OFS-k with and without NMF across different values of $k$. Bold values indicate the better-performing results between the two NMF versions for a given $k$. Red font highlights the overall best-performing results across all $k$ values for the better NMF version.}
\label{table:ofs_k}
\centering
\scriptsize 
\setlength{\tabcolsep}{3pt} 
\begin{tabular}{l | c c | c c | c c | c c}
\hline
\hline
& \multicolumn{2}{c|}{$k=1$} & \multicolumn{2}{c|}{$k=2$} & \multicolumn{2}{c|}{$k=3$} & \multicolumn{2}{c}{$k=4$} \\
& AUC & F1 & AUC & F1 & AUC & F1 & AUC & F1 \\
\hline
OFS-k w/o NMF & 0.748 & 0.627 & 0.756 & 0.633 & 0.758 & 0.633 & 0.760 & 0.640 \\
OFS-k w/ NMF & \textbf{0.954} & \textbf{0.847} & \textbf{0.968} & \textbf{0.867} & \textbf{\textcolor{red}{0.974}} & \textbf{\textcolor{red}{0.870}} & \textbf{0.962} & \textbf{0.866} \\
\hline
\end{tabular}
\vspace{-0.5cm} 
\end{table}

\subsubsection{Evaluations}
\label{eval_ofs}
We also quantitatively evaluated the alignment of our proposed OFS-k metric with human annotations, following the same procedure used for comparing content memorization metrics. As shown in Tab. \ref{table:ofs_k}, OFS-k achieves high classification performance for motion memorization, providing a valuable tool for further research in this area. 
Additionally, we performed an ablation study on the NMF and hyperparameter analysis for \(k\) in OFS-k. From Tab. \ref{table:ofs_k}, we observe that incorporating NMF significantly improves AUC and F1-score, demonstrating its effectiveness in filtering out similar but non-memorized natural motions. This improvement aligns with the more practical notion of motion memorization, emphasizing the importance of privacy preservation. Among the tested values of \(k\), we found that an intermediate value of $k=3$ achieves the best trade-off, yielding the highest AUC and F1-score, whereas $k=2$ achieves the second-best. This supports our earlier intuition that setting \(k\) too low makes the metric overly strict, potentially leading to a high false positive rate by categorizing natural, non-memorized motions as memorized. On the other hand, setting \(k\) too high risks missing cases where only part of the video is memorized, resulting in false negatives.

We also present qualitative results in Fig. \ref{fig:figure5_positive} and \ref{fig:figure5_negative}, where Fig. \ref{fig:figure5_positive} highlights examples of video pairs with high OFS-3 scores. In the first and second blocks, the motions are characterized by a unique and different rotational pattern, with some regions remaining static while others show varying flow magnitudes. These pair reports OFS-3 scores of 0.9012 and 0.9393. In the third block, the plane flies over a horizontal road sign with a similar motion, yielding a score of 0.8523. The final block depicts the motion of a flag, where the logo remains static while the surrounding area flows in a highly similar manner, reporting a score of 0.8806. 
In Fig. \ref{fig:figure5_negative}, we present examples of training and generated video pairs with similar optical flow but representing natural motions like camera panning that do not pose privacy risks. In these cases, the optical flow vectors for each pixel exhibit uniform angles, reflecting consistent motion across the frames. Our proposed NMF effectively filters out such natural motions, ensuring they are not counted as memorized cases.

%% file: sec/4_case_study.tex
\section{Systematic extraction and analysis of memorization in VDMs}
\label{sec:case_study}
\paragraph{Motivations.}
Previous work~\citep{frame_by_familiar_frame} only investigated generated videos sourced from project websites of several VDMs and identified instances of their memorization of the training data. However, the frequency and extent of these memorization issues remain unclear. To address this, we leverage insights into the causes of memorization in diffusion models and apply our proposed metrics for quantifying memorization. This allows us to systematically extract memorized cases and report statistics on the degree and frequency of memorization across multiple VDMs.

In terms of the scope, previous work focused only on memorization issues in unconditional and image-conditional VDMs using small datasets such as UCF-101~\citep{ucf-101} and Kinetics-600~\citep{kinetics-600}. However, for text-to-video (T2V) generation, their study lacked a comprehensive investigation of various VDMs trained on larger datasets like WebVid-10M~\citep{webvid-10m}, which serves as the foundation for many VDMs used in video generation~\citep{make-a-video, modelscope, videocrafter1} and video editing~\citep{sparsectrl, animatediff, moonshot}. To address this gap, we analyzed the WebVid-10M dataset and conducted the first investigation into the memorization behavior of several open-source VDMs~\citep{modelscope, lavie, videocrafter1, videocrafter2} trained on this dataset.

Additionally, their scope only addresses the VDMs' memorization of video training data, while neglecting the memorization of image training data. Specifically, \cite{frame_by_familiar_frame} reported that VDMs extended from pretrained text-to-image (T2I) models can benefit from their inherent capability to generate creative image outputs, which can reduce memorization on video datasets. However, we have demonstrated that this benefit comes at the cost of increased memorization of the image data inherent in the pretrained T2I backbone. This reveals an overlooked trade-off where reducing video dataset memorization leads to the unintended consequence of memorizing image data.

\subsection{Preparing the prompt datasets}
\label{extraction}
Previous studies on memorization in image diffusion models (IDMs)\citep{carlini_2023_usenix, somepalli_2023_neurips, Webster2023} have shown that a significant cause of memorization is the duplication of training data, which leads to overfitting. Leveraging these insights, we begin our analysis by examining the duplication within the widely used WebVid-10M dataset\citep{webvid-10m}. We extract features from the dataset using the Inflated 3D Convolutional Network (I3D)\citep{i3d}, a widely used video feature extractor pretrained on Kinetics\citep{kinetics-600}. I3D extends 2D convolutional networks to 3D, enabling the capture of spatiotemporal features. We compute cosine similarities between training data pairs and identify the most duplicated videos. From this analysis, we extracted the first 500 unique captions corresponding to the most duplicated videos, forming the WebVid-duplication-prompt dataset, which will be used for extracting memorization cases in various video diffusion models (VDMs).

For analyzing VDMs' memorization of image datasets, we utilize the open-sourced prompt datasets provided by \cite{Webster2023}. These datasets organize the most commonly memorized prompts for several IDMs, including Stable Diffusion's memorization of the LAION dataset.

\begin{table}[h]
\footnotesize
\begin{center}
\caption{Quantifying the percentage of memorized generations and average similarity scores for both content and motion memorization across various open-source VDMs, evaluated on both video and image datasets. Content memorization is assessed using GSSCD, while motion memorization is evaluated using OFS-3. Memorization similarity represents the average similarity scores. Regarding the memorization percentage, memorization is determined by thresholds of 0.4 for GSSCD and 0.5 for OFS-3, above which the generation is classified as memorized.}
\label{table:content_motion_comparison}
\begin{tabular}{l|l | c c | c c}
\hline
\hline
Type & VDM & \multicolumn{2}{c|}{Content Mem} & \multicolumn{2}{c}{Motion Mem} \\
& & \%Mem & Sim & \%Mem & Sim \\
\hline
T2V & ModelScope (WebVid) & 18.02 & 0.33 & 33.10 & 0.31 \\
T2V & ModelScope (LAION) & 15.40 & 0.29 & - & - \\
T2V & LaVie (WebVid) & 20.21 & 0.33 & 25.25 & 0.26 \\
T2V & VideoCrafter1 (WebVid) & 3.81 & 0.28 & 19.25 & 0.19 \\
T2V & VideoCrafter2 (WebVid) & 3.93 & 0.27 & 22.68 & 0.19 \\
UV & RaMViD (UCF-101) & 45.00 & 0.43 & 23.75 & 0.09 \\
\hline
\end{tabular}
\end{center}
\end{table}

\subsection{Extracting data from VDMs and analysis}
\label{analysis}
\textbf{Text-conditional generations}. To evaluate text-conditional (T2V) models' memorization of video training data, we conducted experiments on several open-source VDMs trained or partially trained on WebVid-10M, including ModelScope~\citep{modelscope}, LaVie~\citep{lavie}, VideoCrafter1~\citep{videocrafter1}, and VideoCrafter2~\citep{videocrafter2}.
As shown in Tab. \ref{table:content_motion_comparison} and Fig. \ref{fig:figure1}, memorization is widespread across all models using WebVid-10M as part of their training set. Notably, LaVie also utilized the closed-source Vimeo25M dataset, which contains 25 million high-definition, watermark-free text-video pairs, alongside WebVid-10M during training. Intuitively, this should allow the model to generalize better by learning from more diverse sources. However, it still exhibits a 20.21\% memorization rate on WebVid-10M. Due to the closed-source nature of the Vimeo25M dataset, we are unable to investigate its memorization extent, but we believe similar risks apply. This highlights the need for researchers to ensure that even closed-source datasets are free of copyrighted images and videos, as they may pose as legitimate risks as the open-source ones. Similarly, VideoCrafter1 and VideoCrafter2 also report training on private datasets in addition to WebVid-10M. Therefore, the numbers reported in Tab. \ref{table:content_motion_comparison} may underestimate the severity of the memorization issue, as the risk of memorizing content from other datasets remains unaccounted for.

For analyzing the memorization of image training data in T2V models, we experimented with open-source VDMs~\citep{modelscope} that utilize text-to-image (T2I) model backbone of Stable Diffusion~\citep{sd_2022_cvpr}. 
From Tab. \ref{table:content_motion_comparison} and Fig. \ref{fig:figure4}, we observe evidence of ModelScope's generations replicating the image training data. This demonstrates that VDMs are also capable of memorizing image training data, a phenomenon overlooked in the current literature.

\textbf{Unconditional generations}. For evaluating unconditional (UV) models, we analyzed RaMViD~\citep{ramvid} pretrained on UCF-101~\citep{ucf-101}. We found that 45\% of 1,000 generations were memorized instances. As RaMViD was not pretrained on any T2V models, we did not analyze its memorization of image datasets. Qualitative examples can be found in Fig. \ref{fig:figure2}.

%% file: sec/5_strategies.tex
\vspace{-0.4cm}
\section{Remedies}
\label{sec:strategies}
\vspace{-0.2cm}
\subsection{Motivation}
\label{motivation}
\vspace{-0.2cm}
The only previously proposed solution for mitigating memorization in VDMs involves using a text-to-image diffusion model as a backbone, then fine-tuning it on a video dataset to reduce memorization of video data~\citep{frame_by_familiar_frame}. However, as shown in Sec. \ref{sec:case_study}, this approach leads to the unintended consequence of the model memorizing image data, thus reducing its overall effectiveness.

As a preliminary investigation into memorization in VDMs, we draw inspiration from the IDM memorization domain, which offers several potential strategies. One such approach is to retrain VDMs on de-duplicated datasets, as duplication has been identified as a major factor leading to overfitting and memorization. This suggests that future pre-training of VDMs should prioritize training on de-duplicated data or using pre-trained backbones that have been trained on such data. However, two challenges arise: First, duplication is not the only cause of memorization. For example, text captions in different expressions may share similar meanings, which could still trigger memorization when the user prompt is overly specific. Second, retraining on de-duplicated data is computationally expensive and inefficient.
Another approach involves adapting mitigation strategies from the image domain to VDMs, such as adding randomness to input prompts~\citep{somepalli_2023_neurips}, employing prompt engineering~\citep{wen_2024_iclr}, or leveraging guidance strategies during inference~\citep{chen-amg}. However, they typically come at the cost of utility, such as lower quality or reduced text alignment in the generated outputs.

In contrast, we argue that inference-time detection strategies are more practical and impactful. These strategies can effectively and efficiently reduce memorized generations by allowing for early exits during inference upon detecting memorization and advising users to input alternative prompts to prevent legitimate risks.
In terms of effectiveness, inference-time detection strategies do not modify the inference process itself but rely on memory-efficient cache signals to detect memorization, thereby fully preserving output utility. 
In terms of efficiency, these strategies avoid the computational infeasibility of searching through entire training datasets for detection, which is challenging for high-performing models trained on vast amounts of data. Instead, inference-time detection strategies can perform detection during the inference process, which typically takes only tens of seconds. Moreover, our proposed detection method can achieve accurate detection within 1-step and 10-step inference processes that only take seconds.

Thus, developing an efficient inference-time detection strategy for both types of memorization in VDMs could significantly contribute to the field. To this end, we adapt the image memorization detection strategy~\citep{wen_2024_iclr} to the video domain, incorporating the temporal dimension to enable the detection of both content and motion memorization. This serves as a solid foundation for future improvements.

\subsection{Detection strategies for both content and motion memorization}
\label{detection}
In the IDM memorization domain, a recent work~\citep{wen_2024_iclr} introduces a method for memorization detection using the magnitude of text-conditional predictions, expressed as:
\begin{equation}
m_t=\|\epsilon_\theta(x_t, e_p)-\epsilon_\theta(x_t, e_\phi)\|_2
  \label{eq:magitude}
\end{equation}
where \( \epsilon_\theta(x_t, e_p) \) is the denoiser's text-conditional noise prediction, while \( \epsilon_\theta(x_t, e_\phi) \) is its unconditional noise prediction.
This method builds on the observation that, under identical initialization, outputs conditioned on different text prompts tend to exhibit similar semantic properties, resulting in relatively small magnitude values throughout the inference process. However, for prompts \( e_p \) that are prone to memorization, the text condition \( \epsilon_\theta(x_t, e_p) \) consistently steers the generation toward memorized outcomes, regardless of initialization, thereby producing significantly larger magnitude values. Thus, a larger magnitude serves as a signal for potential memorization. 

We extend this detection strategy to VDMs to enable the detection of both content and motion memorization, adapting it to account for the temporal dimension inherent in videos. In IDM, the magnitude is a 3-dimensional tensor with the shape \([C, H, W]\), where \(C\) is the number of channels, and \(H\) and \(W\) represent the height and width of the latent noise prediction \( \epsilon_\theta(x_t, e_p) \) and \( \epsilon_\theta(x_t, e_\phi) \). In VDMs, this expands to a 4-dimensional tensor with the shape \([C, F, H, W]\), where \(F\) is the number of frames in the video.

\vspace{-0.2cm}
\subsubsection{Content Magnitude}

For content magnitude, we follow the same intuition behind our content memorization metric (GSSCD) that emphasizes frame-wise memorization, where frame-wise similarities are computed, and the maximum value is used as the overall score. Similarly, we compute the frame-wise magnitude for each of the \(F\) frames, resulting in \(F\) magnitudes. The maximum magnitude is then taken as the content magnitude:
\begin{equation}
m_{\text{cont}} = \max \{ \|\epsilon_\theta(x_t^i, e_p) - \epsilon_\theta(x_t^i, e_\phi)\|_2 : i = 1, \dots, F \}
\end{equation}
This maintains coherence and simplicity while also effectively detecting content memorization.

\vspace{-0.2cm}
\subsubsection{Motion Magnitude}

For motion magnitude, our approach leverages the temporal dimension by analyzing frame transitions. The frame transition is defined as the difference between each consecutive pair of frames. Given \(F\) frames, this results in \(F-1\) transitions. To compute the motion magnitude, we compare the transitions between the text-conditional prediction and the unconditional prediction:
\begin{equation}
m_{\text{mot}} = \max \{ \|\Delta \epsilon_\theta(x_t^i, e_p) - \Delta \epsilon_\theta(x_t^i, e_\phi)\|_2 : i = 1, \dots, F-1 \}
\end{equation}
where \( \Delta \epsilon_\theta(x_t^i, e_p) = \epsilon_\theta(x_t^{i+1}, e_p) - \epsilon_\theta(x_t^i, e_p) \) represents the $i$-th frame transition at step \(t\). This method captures whether the transitions between frames in the text-conditional generation are ``abnormal" compared to the unconditional prediction, thus signaling motion memorization that overfits the motion in the training videos.

\textbf{Remark on the alternative approach of using optical flow}. We also explored using optical flow to quantify motion transitions. This involved decoding intermediate results of \(\epsilon_\theta(x_t^{i}, e_p)\) and \(\epsilon_\theta(x_t^{i+1}, e_p)\) pairs for each frame \(i\) from each inference step \(t\) back into image space using the diffusion model’s decoder, followed by computing optical flow with RAFT~\citep{raft}. However, this approach yielded unsatisfactory results, as the decoded images were likely out-of-distribution for RAFT compared to the dataset it was trained on. This mismatch led to inaccurate flow estimations, making this method inferior to our proposed approach. Another limitation of using RAFT in this case is the significant additional computational cost, as the RAFT model would need to be invoked multiple times during every inference step. By directly leveraging the latent space through frame transitions, our motion magnitude method remains simple yet effective for detecting motion memorization.
\vspace{-0.2cm}
\begin{table}[h]
\footnotesize
\setlength{\tabcolsep}{4pt} 
\renewcommand{\arraystretch}{1.2} 
\begin{center}
\caption{Comparison of content and motion memorization detection performance at different inference steps, using content and motion magnitude as signals respectively.}
\label{table:content_motion_mem}
\begin{tabular}{l|ccc|ccc}
\hline
\hline
& \multicolumn{3}{c|}{Content Memorization} & \multicolumn{3}{c}{Motion Memorization} \\
Steps & AUC & F1 & Time (s) & AUC & F1 & Time (s) \\
\hline
1-step   & 0.892 & 0.749 & \textbf{0.709} & 0.814 & 0.614 & \textbf{0.709} \\
10-step  & 0.895 & 0.896 & 7.094          & 0.926 & 0.775 & 7.094          \\
all-step & \textbf{0.978} & \textbf{0.904} & 35.470 & \textbf{0.933} & \textbf{0.837} & 35.470 \\
\hline
\end{tabular}
\end{center}
\vspace{-0.6cm} 
\end{table}
\subsubsection{Performance evaluation}
Tab. \ref{table:content_motion_mem} presents a comparative analysis of content and motion memorization detection performance across different inference step strategies, using AUC, F1-score, and inference time as evaluation metrics. The 1-step approach leverages the magnitude computed at the first inference step, offering the advantage of rapid detection with a processing time of just 0.709 seconds while already achieving decent performance.
As the number of inference steps increases, both 10-step and all-step approaches significantly improve detection accuracy. The all-step method, which uses the average magnitude over all 50 inference steps, achieves the highest performance with AUC scores of 0.978 for content memorization and 0.933 for motion memorization and F1-scores of 0.904 and 0.837, respectively. 
The results suggest a trade-off between computational efficiency and detection performance. We argue that the 10-step approach balances efficiency and accuracy, offering notable improvements in detection performance without the full computational burden of the all-step method.

%% file: sec/6_conclusion.tex
\section{Conclusion}
\label{conclusion}
In this paper, we address the critical issue of memorization in video diffusion models (VDMs), a largely overlooked area compared to image diffusion models (IDMs). We introduced new, privacy-focused definitions for both content and motion memorization, as well as tailored evaluation metrics that separately assess these types of memorization in VDMs. 
Armed with such metrics, we are the first to systematically extract training data from large video and image datasets and report quantitative results on the extent of memorization in various open-source VDMs.
Our findings demonstrate that memorization is a widespread issue, not only in video datasets but also in image datasets used to train text-to-image (T2I) backbones, which has been previously underestimated in the literature.
Furthermore, we propose effective remedies by adapting image memorization detection strategies to the video domain. These strategies, incorporating the temporal dimension, offer efficient detection of both content and motion memorization, providing a robust foundation for future improvements in privacy preservation within VDMs, particularly as their usage continues to grow. 
We have also discussed the limitations of our work for future improvements; see Appendix Sec. \ref{sec:future_work} for details.

%% file: sec/acknowledgement.tex
\section{Acknowledgement}
This work was supported in part by the Australian Research Council under Projects DP210101859 and FT230100549.

%% file: sec/X_suppl.tex
\newpage
\appendix

\section{Additional qualitative results}
\begin{figure}[b]
  \centering
  \includegraphics[width=\linewidth]{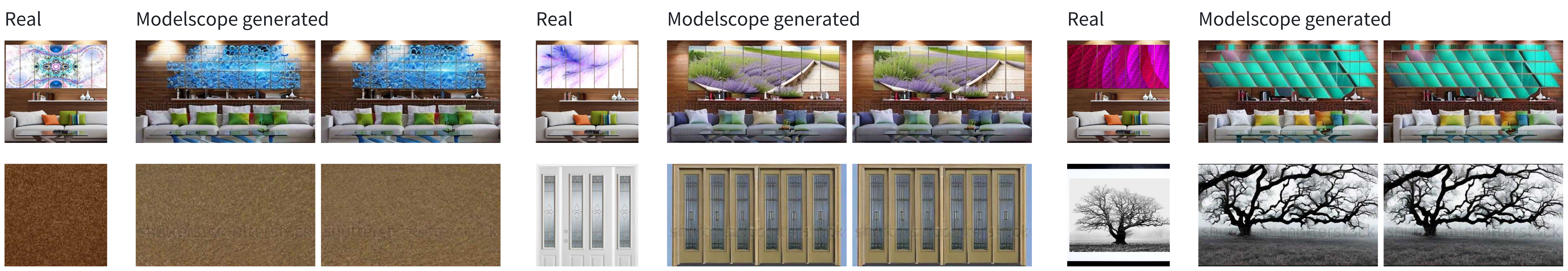}
  \caption{Image training dataset (LAION) being extracted by ModelScopeT2V. For each qualitative example, the left shows images from the training set, while the right displays the most similar frames from the generated videos according to GSSCD.}
  \label{fig:figure4}
\vspace{-0.3cm}
\end{figure}

\begin{figure}[b]
  \centering
  \includegraphics[width=\linewidth]{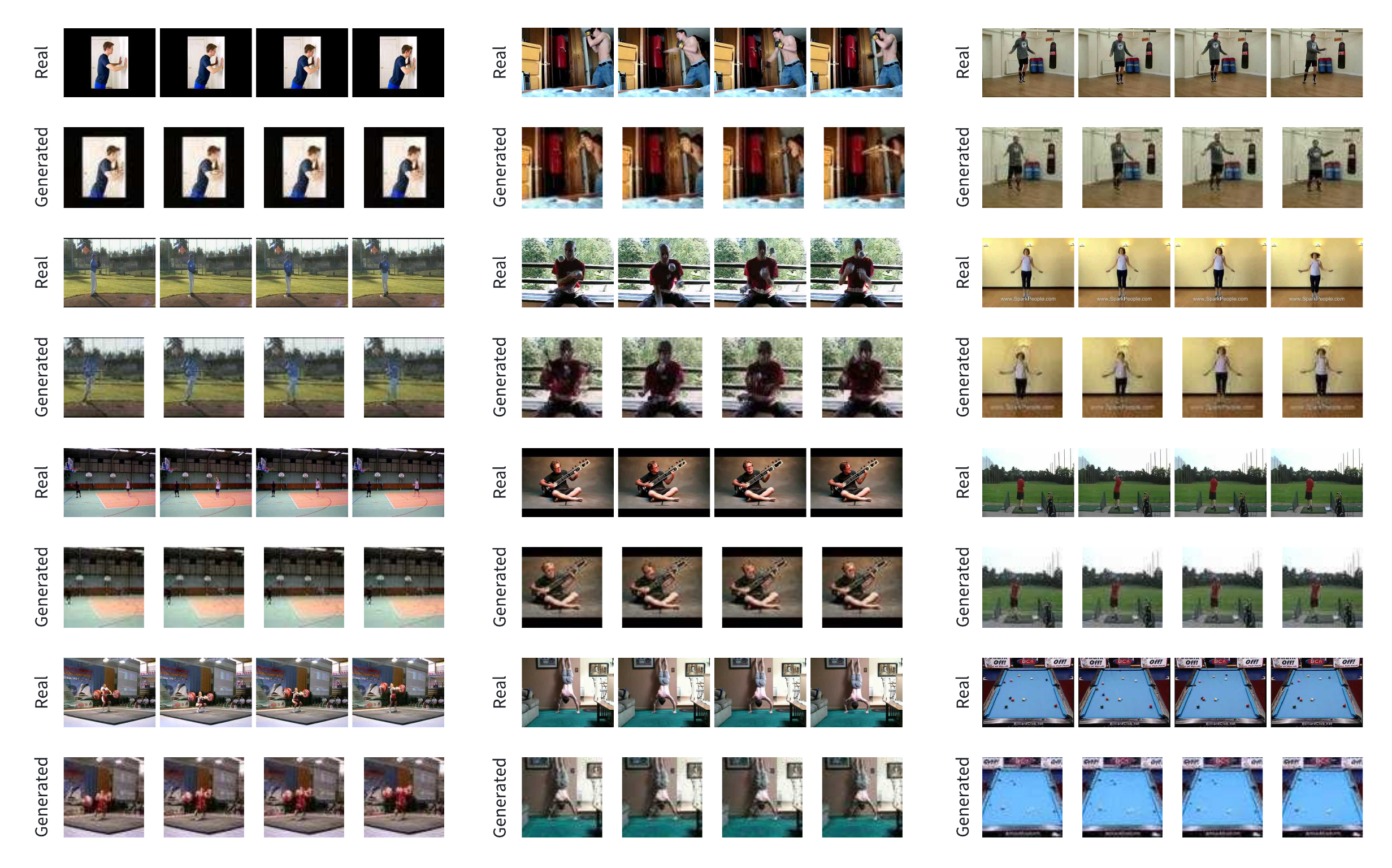}
  \caption{Video training dataset (UCF-101) being extracted by RaMViD's unconditional generation.}
  \label{fig:figure2}
\vspace{-0.3cm}
\end{figure}

\begin{figure}[b]
  \centering
  \includegraphics[width=\linewidth]{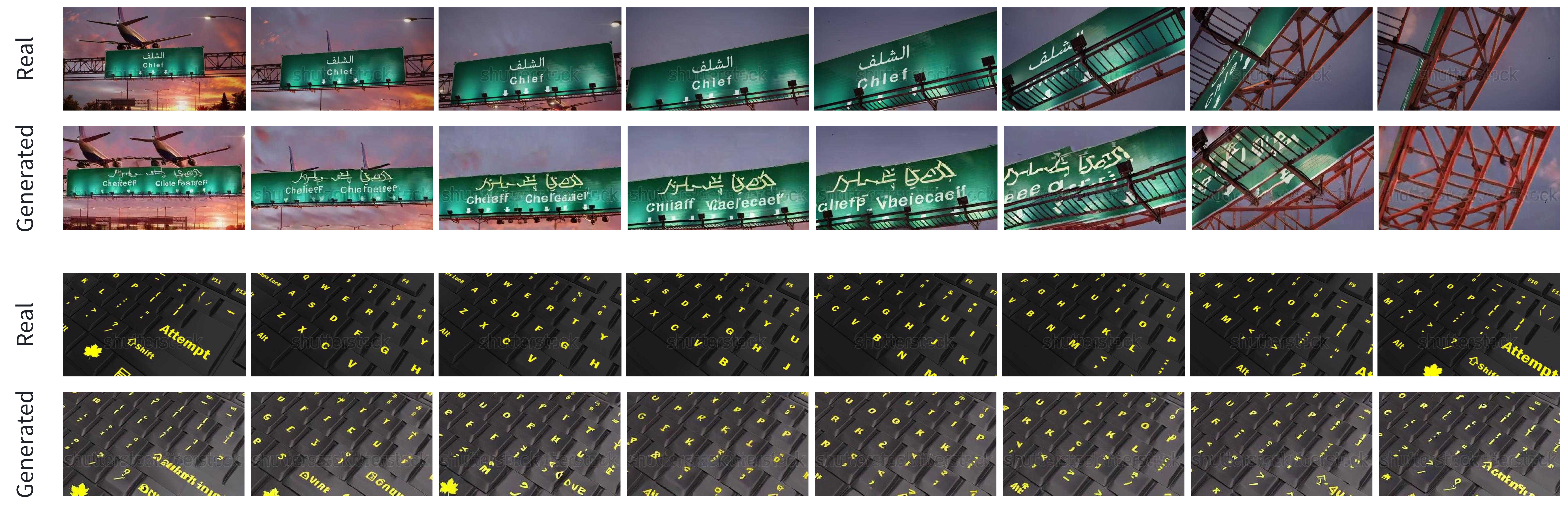}
  \caption{Video training dataset (WebVid-10M) being extracted by ModelScope.}
  \label{fig:figure1_modelscope}
\vspace{-0.3cm}
\end{figure}

\begin{figure}[b]
  \centering
  \includegraphics[width=\linewidth]{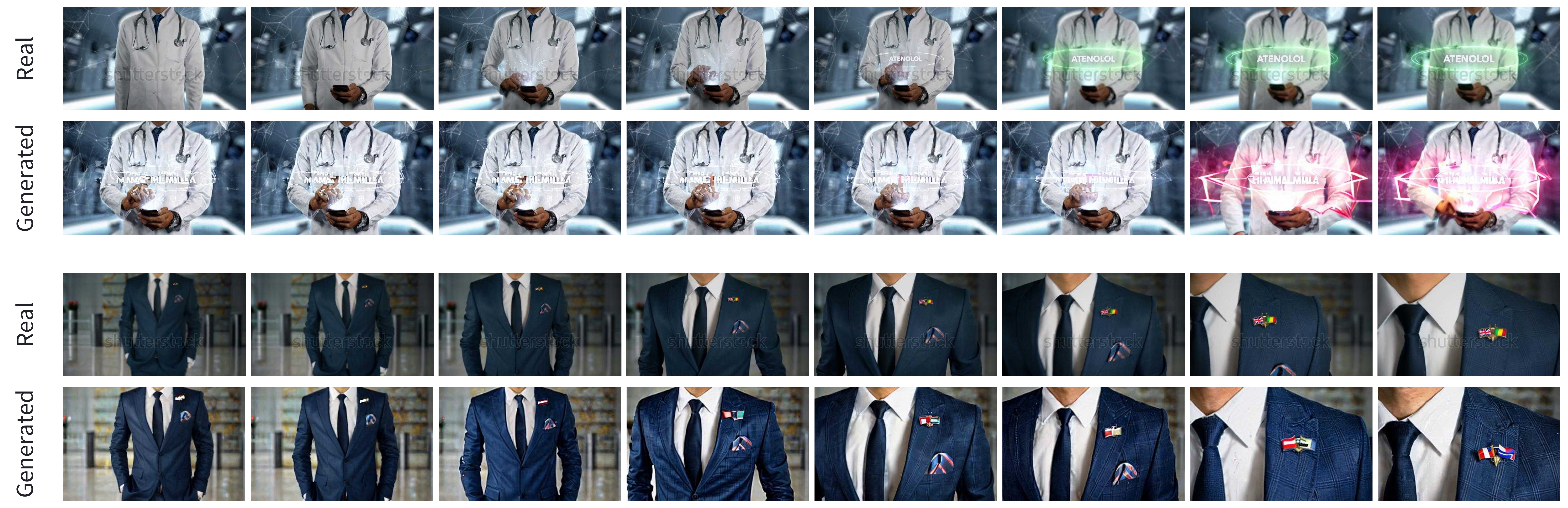}
  \caption{Video training dataset (WebVid-10M) being extracted by LaVie}
  \label{fig:figure1_lavie}
\vspace{-0.3cm}
\end{figure}

\begin{figure}[t]
  \centering
  \includegraphics[width=\linewidth]{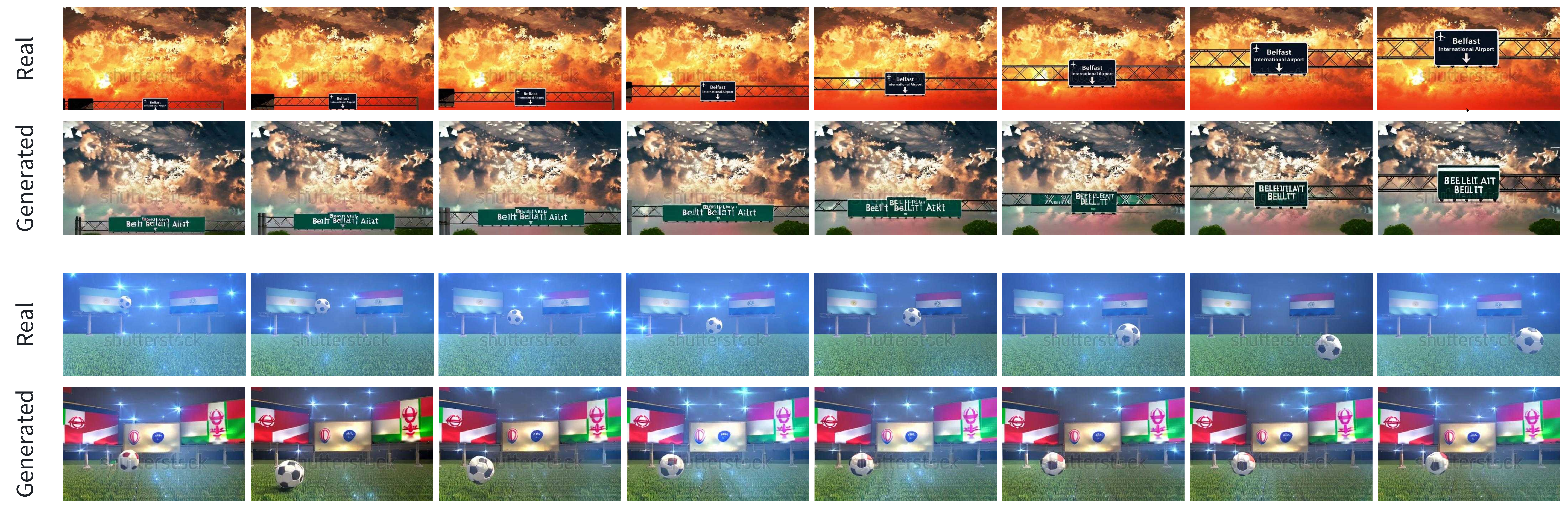}
  \caption{Video training dataset (WebVid-10M) being extracted by VideoCrafter1}
  \label{fig:figure1_videocrafter1}
\vspace{-0.3cm}
\end{figure}

\begin{figure}[t]
  \centering
  \includegraphics[width=\linewidth]{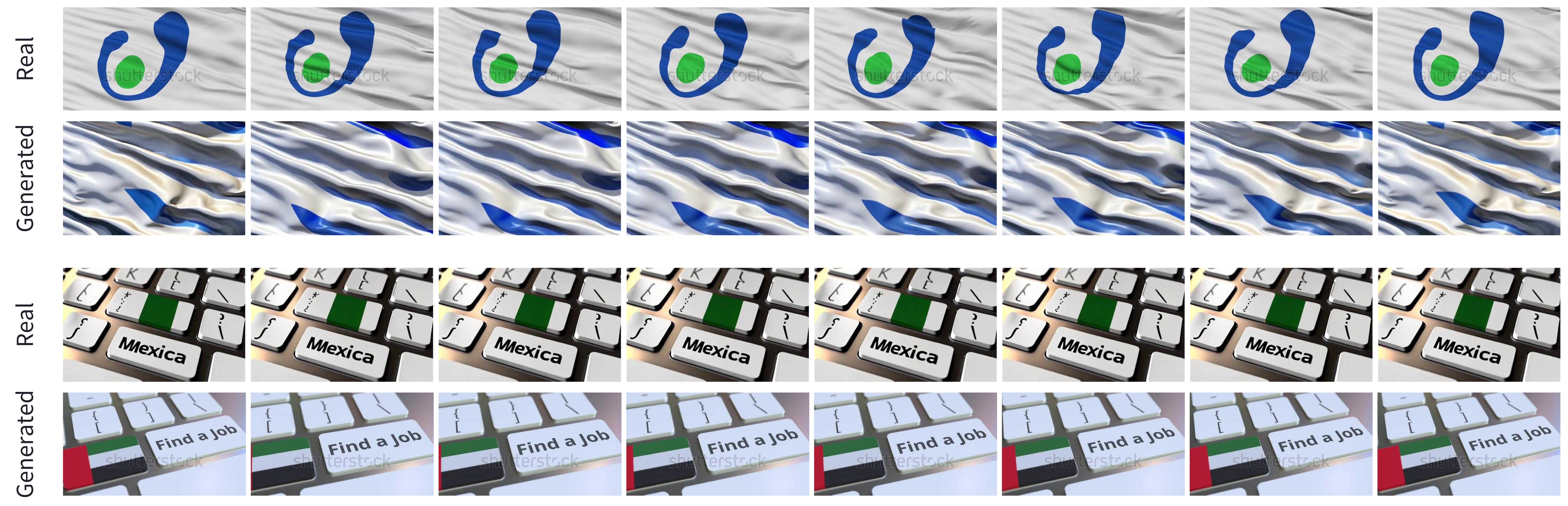}
  \caption{Video training dataset (WebVid-10M) being extracted by VideoCrafter2}
  \label{fig:figure1_videocrafter2}
\vspace{-0.3cm}
\end{figure}

\begin{figure}[t]
  \centering
  \includegraphics[width=\linewidth]{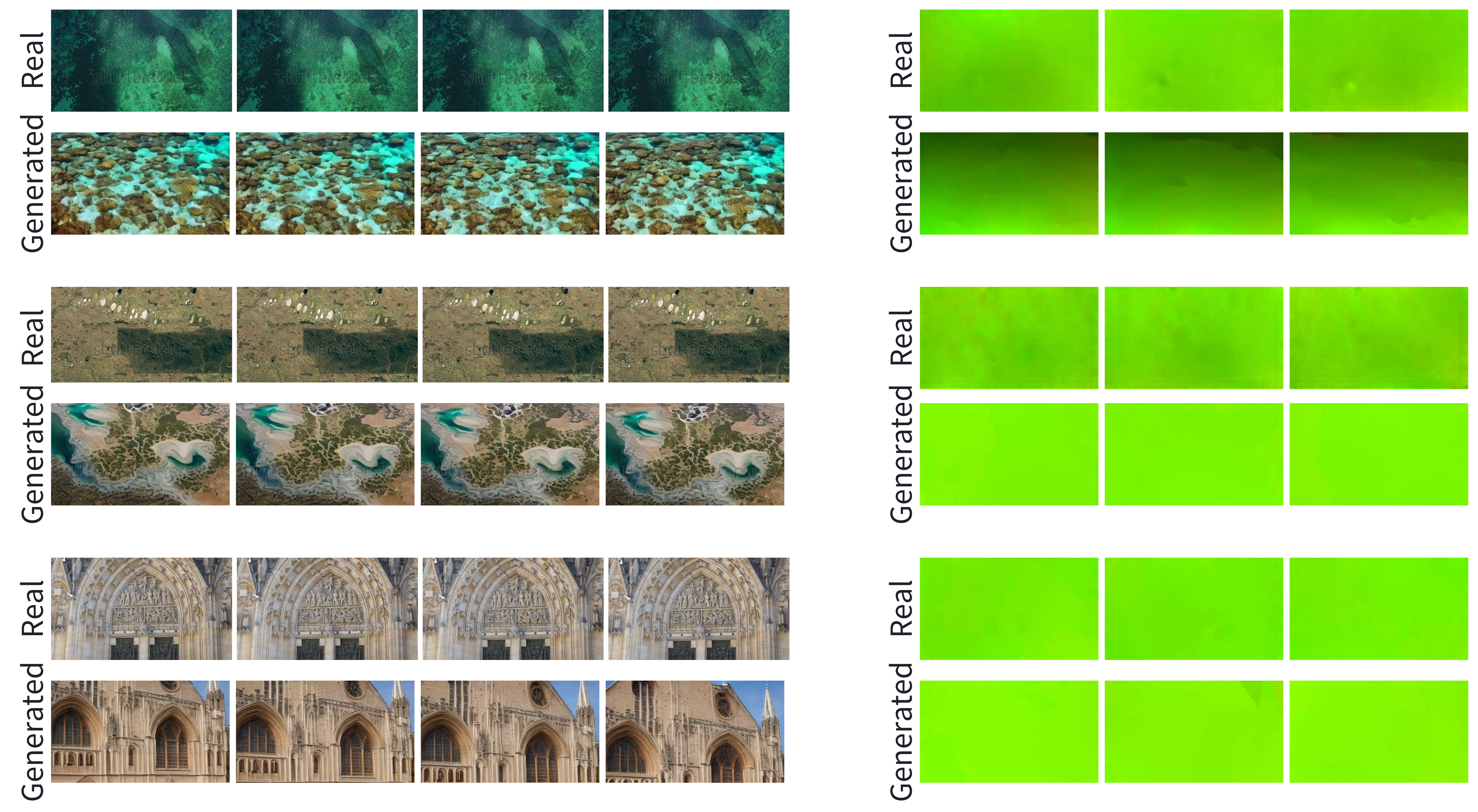}
  \caption{Examples of training and generated video pairs with similar optical flow but representing natural motions like camera panning that do not pose privacy risks. In these cases, the optical flow vectors for each pixel exhibit uniform angles, reflecting consistent motion across the frames. Our proposed NMF effectively filters out such natural motions, ensuring they are not counted as memorized cases. The left side shows consecutive frame pairs from both the training video and the generated video using ModelScope, while the right side visualizes the optical flow between consecutive frames, computed using RAFT.}
  \label{fig:figure5_negative}
\end{figure}

We present the following additional qualitative results:
\begin{itemize}
    \item Fig. \ref{fig:figure4} demonstrates that Video Diffusion Models (VDMs) can also memorize image training data. 
    \item Fig. \ref{fig:figure2} present results of video training dataset (UCF-101) being extracted by RaMViD's unconditional generation.
    \item Fig. \ref{fig:figure1_modelscope}, \ref{fig:figure1_lavie}, \ref{fig:figure1_videocrafter1}, \ref{fig:figure1_videocrafter2} present additional results of video training dataset (WebVid-10M) being extracted by several open-source T2V VDMs. 
    \item Fig. \ref{fig:figure5_negative} present examples of training and generated video pairs with similar optical flow but representing natural motions like camera panning that do not pose privacy risks.
\end{itemize}

\section{Related work}

\textbf{Video diffusion models}.
Video diffusion models (VDMs) extend the diffusion framework, initially developed for image generation, to model the temporal dynamics in videos. Building on the success of image diffusion models (IDMs) such as Denoising Diffusion Probabilistic Models (DDPMs)\citep{ddpm, iddpm_2021_icml} and Stable Diffusion\citep{sd_2022_cvpr}, VDMs tackle the added complexity of maintaining temporal consistency across frames. 
Early works, like Video Diffusion Models~\citep{vdm}, adapted the diffusion process by incorporating 3D convolutional architectures to handle both spatial and temporal dimensions.
Recent VDM advancements such as Make-A-Video~\citep{make-a-video}, which leverage pretrained text-to-image (T2I) backbones like Stable Diffusion~\citep{sd_2022_cvpr}, have pushed the field forward, expanding the application of VDMs to conditioned video generation from text prompts. Other models, such as Runway Gen-1~\citep{gen-1}, also push boundaries by incorporating text, images, and videos as conditional inputs to better control generated content.
The popularity of VDMs has further surged with advancements such as Sora~\citep{sora}, Gen-3~\citep{gen3}, and discord-based servers~\citep{pika}, which have garnered significant attention on social media. 
Despite this increasing interest, the phenomenon of memorization in VDMs remains largely underexplored.

\textbf{Memorization in diffusion models}.
Memorization in image diffusion models (IDMs) has been extensively studied. Pioneer works such as \cite{carlini_2023_usenix} and \cite{somepalli_2023_cvpr} successfully extracted training data from IDMs, including DDPMs and Stable Diffusion, using datasets like CIFAR-10~\citep{cifar10} and LAION~\citep{LAION5B}. This motivates the subsequent works~\citep{somepalli_2023_neurips, ambient_diffusion, onmem, wen_2024_iclr, chen-amg, ren2024unveiling} to work on analyzing the causes and proposing remedies.
Recently, \cite{frame_by_familiar_frame} explored memorization in VDMs, but the study's definitions and evaluations of content and motion memorization are limited in scope and lack broad applicability. Additionally, their analysis was based on individual instances of generated videos extracted from project websites, rather than a systematic study of training data and the extent of memorization. Furthermore, the study did not explore memorization in text-to-video (T2V) models trained on widely used datasets like WebVid-10M.

In this work, we address these limitations by conducting a thorough investigation on several open-source VDMs, quantifying the percentage of memorized generations and average similarity scores for both content and motion memorization across both video and image datasets.

\newpage
\section{Limitations and future work}
\label{sec:future_work}
Despite its demonstrated effectiveness, our study has limitations that future research can address. 

\textit{Firstly}, our proposed GSSCD builds upon the success of SSCD, the standard for evaluating memorization in IDMs, by extending it to support both image and video memorization in a generalized framework. While this generalization brings practical benefits, GSSCD inherits any limitations inherent in SSCD, especially considering the rapidly evolving field of memorization in IDMs. However, our results validate GSSCD as a simple yet effective adaptation of SSCD, and its design ensures compatibility with future advancements in SSCD, allowing it to benefit from subsequent improvements. Besides, while the design of GSSCD may appear straightforward, it is the result of a deliberate analysis of the shortcomings in existing VDM memorization research. Prior metrics lacked a frame-level focus and failed to account for practical scenarios where frame-level memorization poses privacy risks. GSSCD addresses this gap by introducing a privacy-centered, frame-level definition of memorization in VDMs, making it more meaningful and actionable in real-world applications. Future work could build on this foundation to explore alternative metrics.

\textit{Secondly}, our study focuses on VDMs that have open-sourced both their models and the datasets used for training, as these are essential for analyzing memorization issues comprehensively. Nonetheless, our findings indicate that memorization is a pervasive issue across all tested VDMs, suggesting its likely presence in other models as well.
Importantly, this limitation does not detract from the broader value of our study, as the metrics and detection strategies we propose for content and motion memorization are model-agnostic and applicable to any VDM or IDM leveraging classifier-free guidance (CFG). Our metrics evaluate outputs directly, while the detection strategies rely only on the text-conditional and unconditional noise prediction terms provided by CFG. This ensures wide applicability across various architectures, making our contributions relevant to both existing and future diffusion models.